\documentclass[11pt]{article}

\usepackage[sort,numbers]{natbib}
\usepackage{fullpage}

\usepackage[utf8]{inputenc} % allow utf-8 input
\usepackage[T1]{fontenc}    % use 8-bit T1 fonts
\usepackage{hyperref}       % hyperlinks
\usepackage{url}            % simple URL typesetting
\usepackage{booktabs}       % professional-quality tables
\usepackage{amsfonts}       % blackboard math symbols
\usepackage{nicefrac}       % compact symbols for 1/2, etc.
\usepackage{microtype}      % microtypography
\usepackage{xcolor}         % colors
\usepackage{graphicx}
\usepackage{caption}
\usepackage{subcaption}
\usepackage{amsmath}
\usepackage{amssymb}
\usepackage{appendix}

\setcitestyle{square}
\setcitestyle{numbers}

\title{Fairer and More Accurate Tabular Models Through NAS}

\author{Richeek Das, Samuel Dooley\footnote{Correspondence to: \texttt{samuel@abacus.ai}.
} \\
    Abacus.AI
}

\date{}

\begin{document}

\maketitle

\begin{abstract}
  Making models algorithmically fairer in tabular data has been long studied, with techniques typically oriented towards fixes which usually take a neural model with an undesirable outcome and make changes to how the data are ingested, what the model weights are, or how outputs are processed. 
  We employ an emergent and different strategy where we consider updating the model's architecture and training hyperparameters to find an entirely new model with better outcomes from the beginning of the debiasing procedure. 
  In this work, we propose using multi-objective Neural Architecture Search (NAS) and Hyperparameter Optimization (HPO) in the first application to the very challenging domain of tabular data. 
  We conduct extensive exploration of architectural and hyperparameter spaces (MLP, ResNet, and FT-Transformer) across diverse datasets, demonstrating the dependence of accuracy and fairness metrics of model predictions on hyperparameter combinations. 
  We show that models optimized solely for accuracy with NAS often fail to inherently address fairness concerns. 
  We propose a novel approach that jointly optimizes architectural and training hyperparameters in a multi-objective constraint of both accuracy and fairness. 
  We produce architectures that consistently Pareto dominate state-of-the-art bias mitigation methods either in fairness, accuracy or both, all of this while being Pareto-optimal over hyperparameters achieved through single-objective (accuracy) optimization runs. This research underscores the promise of automating fairness and accuracy optimization in deep learning models.

\end{abstract}

\section{Introduction}
\label{sec:intro}

Tabular data is widespread throughout a wide array of real-world applications, spanning medical diagnosis, housing price prediction , loan approval,
and robotics.
Many of the most sensitive forms of tabular data in real-world applications involve an element of fairness considerations where these tables, either directly or in conjunction, contain a set of protected attributes that partition the dataset into groups where some protected attributes have higher performance than others. \cite{misc_adult_2, ding2021retiring, misc_statlog_(german_credit_data)_144, misc_communities_and_crime_183, blacksvscrime}. Examples of protected attributes include but are not limited to race, gender, age, caste, and religion. Many seemingly unrelated features can build up a sensitive feature, and these are widely application-specific. Unfortunately, both classical and deep learning algorithms, when trained with standard training procedures, i.e., without additional fairness interventions, are generally observed to learn unwanted biases that place privileged groups at a systemic advantage. Further, because of correlated biases, simply removing the protected attribute columns from the training data set does not fix the issue of biased predictions. 

To this end, several debiasing techniques have been proposed, and these intervene at intermediate stages of the machine learning pipeline -- \textbf{pre-processing:} modifying the training data to curb inherent biases, \textbf{in-processing:} building specific classifiers that learn fair representations and reduce bias in predictions, \textbf{post-processing:} modifying the output logits to make them less biased towards the unprivileged label of the protected attribute. Although many of these existing techniques are pretty effective at producing fair predictions, these come at a certain additional computational cost, and the overall pipeline generally takes a hit in prediction accuracy.

None of these approaches take a look at leveraging the implicit bias of a model's \emph{architecture} in order to achieve more desirable, and/or more fair outcomes. 
Neural Architecture Search (NAS) has emerged as a prominent technique for automating the design and optimization of neural network architectures \cite{zoph2017neural}. Though it has managed to search for architectures with superior performance in terms of accuracy, NAS, to the best of our knowledge, has never been widely used to search for architectures that are inherently fair, with only one other application of NAS to fairness topics~\cite{dooley2022importance,dooley2023importance} though in a much simpler domain. This paper delves into the realm of utilizing neural architecture search (NAS) and hyper-parameter optimization (HPO) in tabular data to answer two important research questions (RQs):

\begin{enumerate}
    \item[\textbf{[RQ1]}] Does searching for novel architectures (and hyper-parameters) in a multi-objective setting targeting both accuracy and fairness outperform existing bias mitigation techniques in tabular data?

    \item[\textbf{[RQ2]}] Does chaining existing pre-processing bias mitigation techniques with models found by NAS + HPO (single-objective or multi-objective) help in tackling the accuracy hit taken by the bias mitigation intervention?
\end{enumerate}

In this work, we take a step towards answering these research questions and discover two main research contributions.

\textbf{Our main contributions in this paper are summarized below:}
\begin{itemize}
    \item We try out a wide range of architectural and training hyperparameters on multiple search spaces, including MLP, ResNet, and FT-Transformer. We observe significant variation and tradeoffs in the accuracy and fairness of the model predictions with changes in hyperparameters, revealing that certain subspaces of the search landscape are inherently fairer, more accurate, or both.

    \item We exploit the aforementioned observation and propose a joint optimization of both the architectural and training hyperparameters based on the multi-objective accuracy and fairness response on the validation set. With this method, we discover a set of architectures that are Pareto-optimal over all the existing state-of-the-art bias mitigation techniques in terms of accuracy and multiple fairness metrics. We also observe that these architectures are vastly fairer (while being equally accurate) than a single-objective neural architecture optimization based on just the accuracy response.
\end{itemize}

\section{Related Work}

\paragraph{Fairness in Tabular Data}

The use of tabular data has been the cornerstone of core machine learning advancements for years, with a steady increasing interest in fairness considerations in the last decade.
Fairness in machine learning has been studied extensively with various styles of approaches to such research including: observational studies~\cite{buolamwini2018gendershades, dooley2022robustness, cherepanova2023deep, dooley2021comparing}, 
interventions~\cite{friedler2019comparative, raghavan2020mitigating, }, 
field studies~\cite{liu2023reimagining, cherepanova2023deep, }, 
theoretical work~\cite{dwork2012fairness, dwork2018group, }, 
and debiasing techniques. 
In this work, we propose a new style of debiasing work which does not fall within the established categories in fairness in tabular data. 
Typically, there are one of three categories for debiasing algorithms in tabular data: pre-processing \citep[e.g.,][]{Feldman2015Certifying, ryu2018inclusivefacenet, quadrianto2019discovering, wang2020mitigating},
in-processing \citep[e.g.,][]{zafar2017aistats, zafar2019jmlr, donini2018empirical, goel2018non, wang2020mitigating,nanda2021fairness}, or post-processing \citep[e.g.,][]{hardt2016equality,wang2020fairness}.
However, in this work, we employ a different approach to find novel architectures at the beginning of the training pipeline, where  prior work in fairness approaches in tabular data make interventions to models and systems without changing architectural elements of a model.

\paragraph{Neural Architecture Search (NAS) and Hyperparameter Optimization (HPO).}
The common approach to feature engineering in deep learning is to use manually-designed feature extractors, which requires the designer to have a lot of intuition about the problem and past successful approaches in order to achieve success.
On the other hand, approaches like Neural architecture search (NAS) \citep{elsken2019neural}
try to automate these feature extractors by automatically desingning the architectural network for a given task.
This is even a subset of the larger field of Hyper Parameter Optimization (HPO) \citep{feurer2019hyperparameter}, which can view architectural details as special cases of more general hyperparameters traditionally associated with deep learning, like the batch size, loss function, learning rat, optmizer, dropout, etc.
The field of NAS has seen widespread success in various fields in visual applications, like image classification and object detection~ \citep{liu2018darts,zela2019understanding,xu2019pc, pham2018efficient,cai2018proxylessnas}. 
Often, NAS or HPO methods are primarily focused on a singular objective: accuracy. 
The methods work by searching through a hyperparameter space to find models which maximize the accuracy of the entire system. 
There are some NAS and HPO methods that have also been applied for multi-objective optimization \citep{guo2020single,cai2019once}, e.g., optimizing accuracy and size.
Most relevant to this work is the the approach of~\cite{dooley2022importance,dooley2023importance} which takes a similar NAS+HPO strategy to find fairer facial recognition models through a multi-objective approach.
This work however looks at NAS and HPO in the visual field, where NAS has a proven track record of success, whereas we explore NAS and HPO in the tabular setting.

\paragraph{Biased Tabular Data with NAS and HPO}

Within the realm of tabular data, bias has been addressed with only hyperparameter optimization in only a select few number of works, with no known work using architecture search and hyperparameter optimization jointly.
First, \citet{perrone2021fair} uses a framework with Bayesian optimization as a means of finding highly performant models which also satisfy a bias or fairness constraint.
The popular work HPO algorthim called Hyperband \citep{li2017hyperband} has been extended in the work of \citet{schmucker2020multi} and \citet{cruz2020bandit} to apply to fairness appplications in the multi-objective setting, with a further extension of \citet{schmucker2020multi} into an asynchronous setting done by\citep{schmucker2021multi}.
To the best of our knowledge, no prior work uses any joint neural architecture search and hyperparameter optimization to design fair tabular data models, and only \cite{dooley2022importance,dooley2023importance} uses NAS to design architectures for any fairness application though importantly not in the harder domain of tabular data.

\section{Methods}

This section explains how we employ joint NAS+HPO optimization to find Pareto-optimal/dominant sets of model hyperparameters on the accuracy and fairness front. Firstly, we review our search space designs and evaluation metrics for our analyses. Then, we propose our multi-objective search strategy in Section~\ref{ssec:optstrat} and provide a detailed set of empirical results answering our aforementioned research questions in Section~\ref{sec:results}.

\subsection{Search Space Design}

We use three model classes, namely MLP (Multi-layer Perceptron), ResNet, and FT-Transformer, as our base models for the search spaces since these are the most common and highest performing architectures in tabular data at this time. We design these search spaces on top of the \texttt{rtdl} package (introduced by \cite{gorishniy2021revisiting}) for easy reproducability. The hyperparameter search space is more or less similar for each of them, while the architecture-specific parameters differ. 

\paragraph{Training Hyperparameter Search Space.} We optimize for the learning rate, weight decay, batch sizes, and dropout rates for each of the aforementioned search spaces as a part of the HPO routine.

\paragraph{Architectural Hyperparameter Search Space.} For MLP and ResNet, we control the number of trainable layers and the count of neurons in each layer. For the FT-Transformer search space, we optimize for the number of trainable attention blocks, attention heads, and the count of hidden linear layers and their dimensions. The following table summarizes the number of architectural combinations possible for each of the search spaces:

\begin{table}[h]
\centering
\begin{tabular}{cc}
\toprule
Search Space & Combinations \\
\midrule
MLP & 875 \\
ResNet & 350 \\
FT-Transformer & 324 \\
\bottomrule
\end{tabular}
\end{table}

Note that training hyperparameters like learning rates, weight decay, and dropout rates are continuous variables, and combined with train/test batch sizes, they have an infinite number of possible combinations.

As we will notice in Section~\ref{sec:results}, the defined ResNet and FT-Transformer search spaces consistently show a strong tradeoff between accuracy and fairness metrics on the test datasets. However, the MLP space does not exhibit significant potential in achieving inherent fairness-accuracy tradeoff with the change in input hyperparameters -- often converging to the trivial solution for a heavily imbalanced dataset. Hence, we limit our detailed experimental validations in Section~\ref{sec:results} to ResNet and FT-Transformer models.

\subsection{Evaluation, Metrics, and Experimental Design}
\label{ssec:expdesign}

We train and evaluate our model configurations on three benchmark datasets: (1) Adult Income Dataset \cite{misc_adult_2}, (2) COMPAS \cite{misc_communities_and_crime_183}, and (3) Folktables ACS-Income task (for Pennsylvania 2018) \cite{ding2021retiring}. We choose the Adult and COMPAS datasets as they are widely used within the fairness in tabular data community, while also including Folktables as a more modern update to the traditional datasets. 

A full training of each model comprises 10 epochs, and we record a set of standard accuracy and fairness metrics -- that are well-defined for tabular binary classification tasks. Because of the imbalanced nature of the data set with more focus on the negative predictions (unfair outcomes), we chose balanced accuracy $(0.5\times(\text{Sensitivity} + \text{Specificity}))$ and standard accuracy as the set of performance measures. To evaluate the prediction fairness, we use the following standard binary fairness metrics: disparate impact, statistical parity difference, average odds difference, and equal opportunity difference 

We use as baselines for our method many existing de-biasing methods in tabular data. These baselines include state-of-the-art bias mitigation techniques and a set of off-the-shelf models like Logistic Regression, MLP, ResNet, and FT-Transformer with default hyperparameters; we call these naive baselines. For the pre-processing bias mitigation intervention, we have Reweighing \cite{reweighing} with Logistic Regression and each of Disparate Impact Remover \cite{disparateimpactremover}, Learning Fair Representations \cite{lfr}, and Optimized Preprocessing \cite{optimpreproc} with Logistic Regression, MLP, ResNet, and FT-Transformer learning methods. For the in-processing baselines, we plot results for Adversarial Debiasing \cite{adversarialdebiasing}, Gerry Fair Classifier (that is a modification of the FairFictPlay algorithm mentioned in \cite{gerryfair1, gerryfair2}), Prejudice Remover (based on the Kamishima algorithm of \cite{prejudiceremover}), Exponentiated Gradient Reduction \cite{expgradreduc}, and Grid Search Reduction \cite{gridsearchreduc} in their default settings. Note that for the in-processing methods, we do not change the base model used as the classifier; for instance, the logistic regression classifier in Adversarial Debiasing as proposed in the original publication \cite{adversarialdebiasing} is kept untouched in our baseline runs. Finally, for the post-processing bias mitigation baselines, we compare our results with Calibrated Equality of Odds \cite{calibratedeqodds}, Equality of Odds \cite{eqodds}, and Reject Option Classification \cite{rejectopclas} where each technique acts upon the classifier scores and logits of the aforementioned set of 4 models. We treat these as the set of core baselines for the rest of the paper.

\subsection{Optimization Strategies}
\label{ssec:optstrat}

In our proposed strategies tackling the pre-defined search spaces, we employ multi-fidelity and multi-objective black-box optimization techniques. For the entire black-box optimization, we use the \texttt{SMAC3} package \cite{smac3} wherein the main part consists of Bayesian Optimization and intensification techniques to efficiently decide which of the pairwise configurations performs better. Now, owing to the large number of combinations for architectural and training hyperparameters, multi-fidelity optimization is a necessity in our setup. For instance, our MLP search space mandates 875 different architectural hyperparameter combinations and possibly an infinite set of training hyperparameter combinations owing to the continuous variables. The multi-fidelity approach makes it possible to prematurely evaluate the hyperparameters by optimizing the said combination over a smaller number of epochs or a subset of the dataset. In our experiments, we use Hyperband \cite{hyperband} as the intensifier algorithm for multi-fidelity with the min and max budgets set to 1 and 10 epochs, respectively, and the successive halving $\eta$ to 3.

\paragraph{Multi-objective Optimization.} To answer our research questions in Section~\ref{sec:intro}, we need a method to perform multi-objective optimization targeting accuracy and fairness. This is generally computationally or temporally expensive in a traditional grid or random search methodology. To make the problem tractable, we target improving the worst-case performance by reducing the multiple objectives into a single scalar value. We try out two optimization strategies, namely the weighted mean-aggregation strategy, which is self-explanatory, and ParEGO \cite{parego}, which is a multi-objective extension of the efficient global optimization algorithm proposed in \cite{jones_ego}. In this multi-objective setting, we jointly optimize both accuracy and fairness. We outline a set of results in Section~\ref{sec:results} where each of the aforementioned models is black-box optimized on balanced accuracy and once for each of the following fairness metrics: average odds difference, statistical parity difference, and equal opportunity difference. We note that while optimizing disparate impact as the fairness metric, we encounter numerical instability owing to the ratio of conditional probabilities of imbalanced classes. Further, for a fairer comparison, we perform a single-objective optimization on the standard accuracy metric and observe that the obtained Pareto front produces highly detrimental models to unprivileged classes. These single-objective optimized models offer little to no bias mitigation and, while being very accurate, are highly stereotypical and encompass several harmful biases that directly or indirectly utilize the protected attributes. We talk about these results in more detail in the following section.

\begin{figure}[t!]
\centering
\begin{subfigure}{.33\textwidth}
  \centering
  \includegraphics[width=1\linewidth]{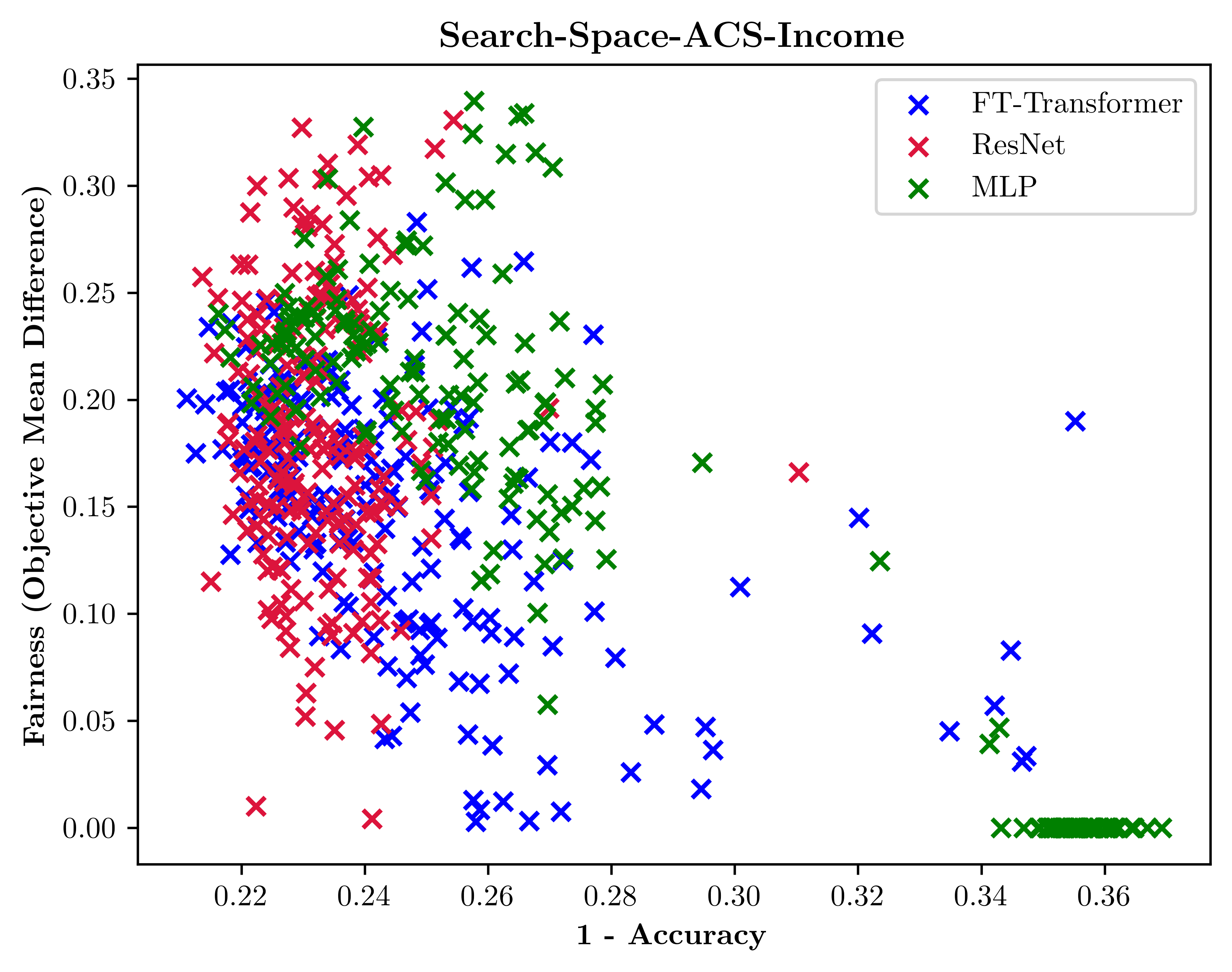}
  % \caption{A subfigure}
  % \label{fig:sub1}
\end{subfigure}%
\begin{subfigure}{.33\textwidth}
  \centering
  \includegraphics[width=1\linewidth]{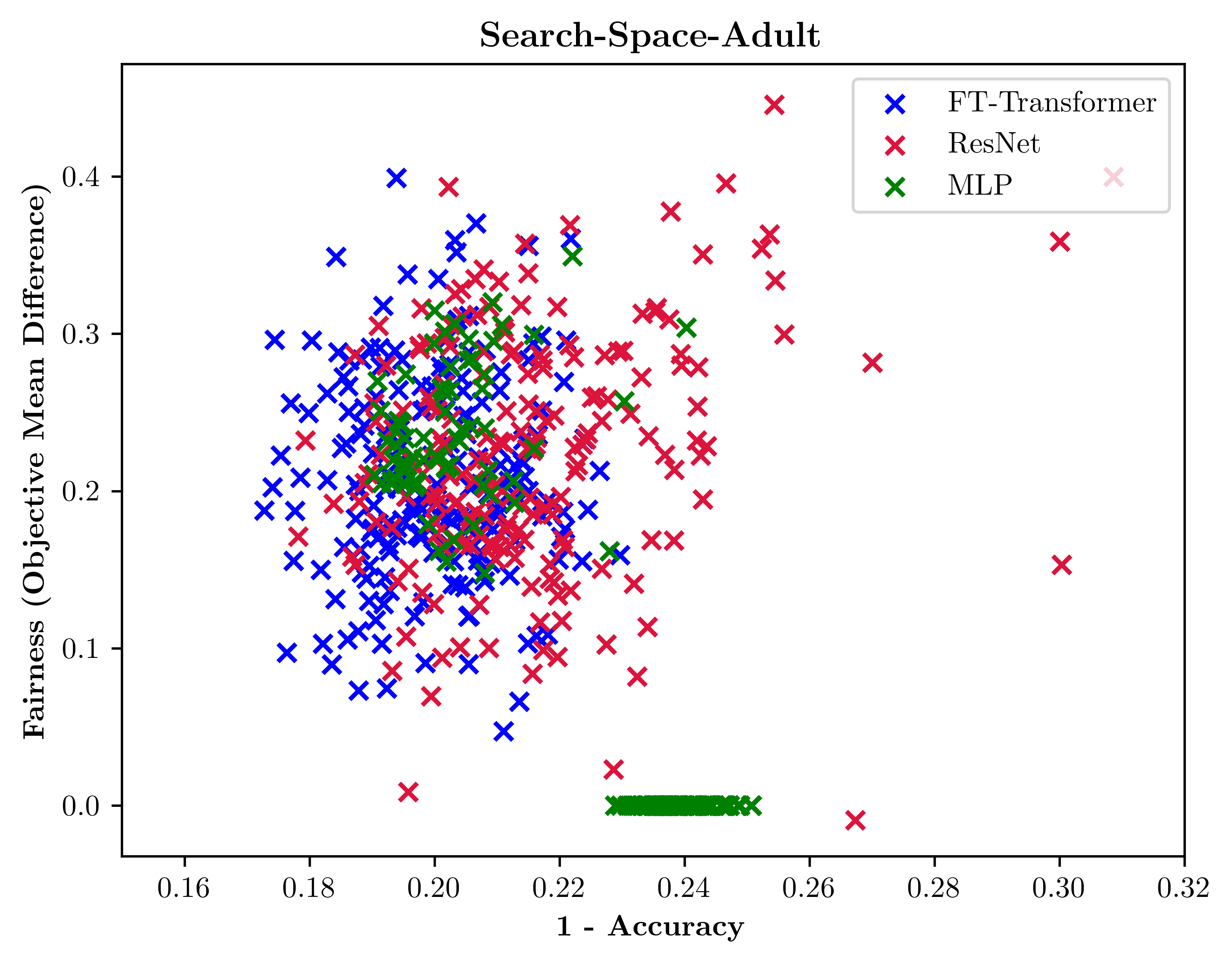}
  % \caption{A subfigure}
  % \label{fig:sub2}
\end{subfigure}%
\begin{subfigure}{.33\textwidth}
  \centering
  \includegraphics[width=1\linewidth]{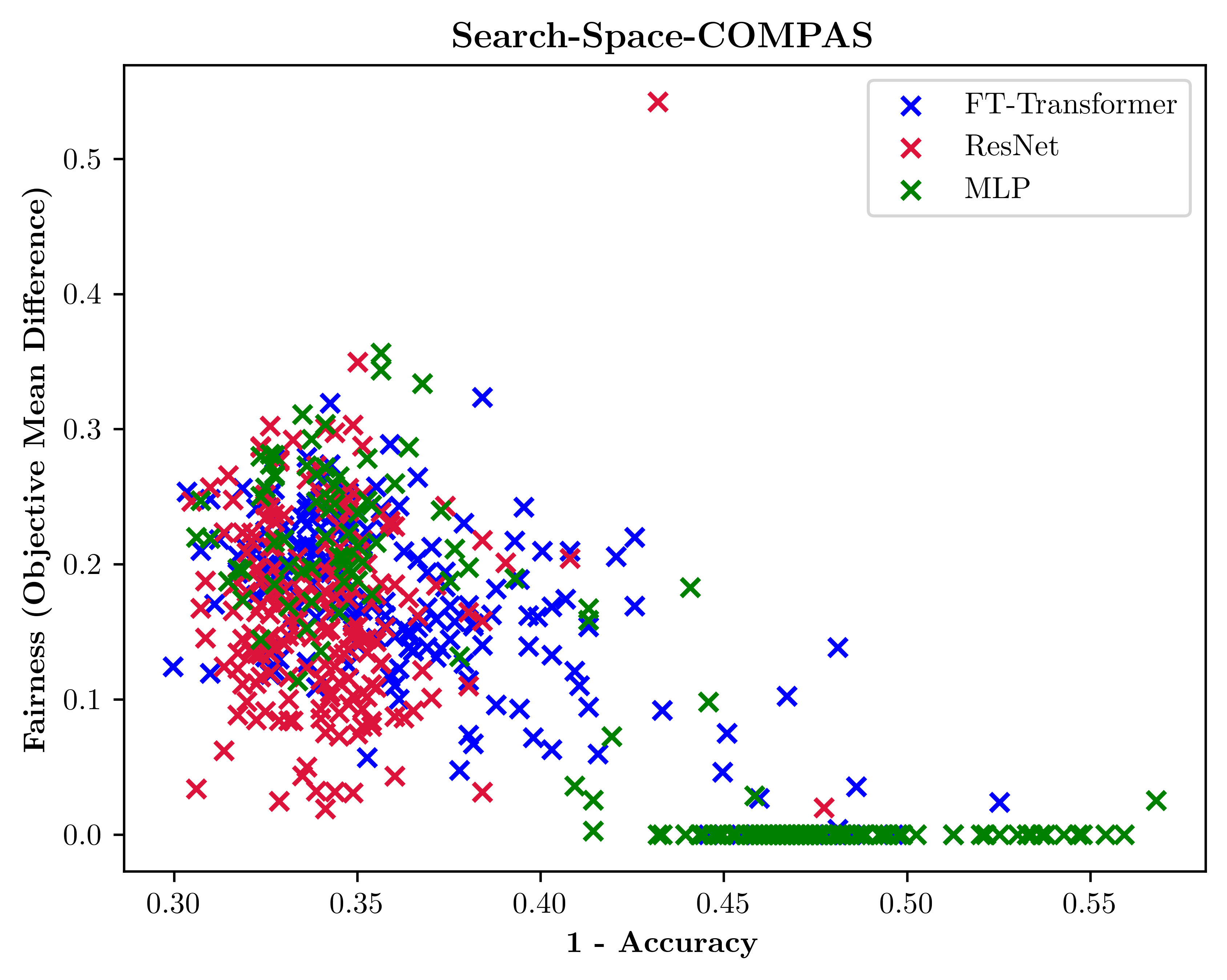}
  % \caption{A subfigure}
  % \label{fig:sub2}
\end{subfigure}
\caption{Search Space Probing for each MLP, ResNet, and FT-Transformer class. We plot the fairness value, here the absolute value of the Mean Difference between the privileged and unprivileged groups versus the accuracy achieved in the predictive task labeled above. We choose a random subset of 100 hyperparameter combinations for each search space. In each of the above scenarios, we see a standard deviation $> 0.051$ for the absolute mean difference and $> 0.023$ for the predictive accuracy. Some hyperparameter choices encounter up to a $\mathbf{0.40}$ shift on the fairness front in each of the benchmarks. \textbf{Note:} The hyperparameter combinations are evaluated and averaged across $6$ seeds.}
\label{fig:sspace}
\end{figure}

\section{Results}
\label{sec:results}

\paragraph{Search Space Probing.} First, we seek to understand the complexity inherent in  designed hyperparameter search spaces entail. We evaluate a subset of all possible model combinations on each of our benchmark datasets and show the dependence of the accuracy and fairness of a trained model on its input hyperparameters. In Figure~\ref{fig:sspace}, we plot the statistical parity difference and the standard accuracy for each sampled hyperparameter combination. Even a simple search space design like ours brings in significant variance in the model's inherent fairness and predictive performance. For instance, the FT-Transformer search space has a standard deviation of $\pm 0.057$ (in mean difference) and $\pm 0.024$ (in err0r) on the ACS-Income dataset. The MLP, ResNet, and FT-Transformer search spaces continue to show these strong trends for all three of our benchmarks -- providing empirical evidence for the inherent fairness of models on tabular data solely based on the hyperparameter choices. However, MLP search space shows a unique tendency to converge to the trivial solution for a heavily imbalanced dataset. A lot of the MLP combinations end up trivial accuracy. Thus, for further experiments, we limit our analyses to only FT-Transformer and ResNet search spaces.

\begin{figure}[t!]
\centering
\begin{subfigure}{1\textwidth}
  \centering
  \includegraphics[width=1\linewidth]{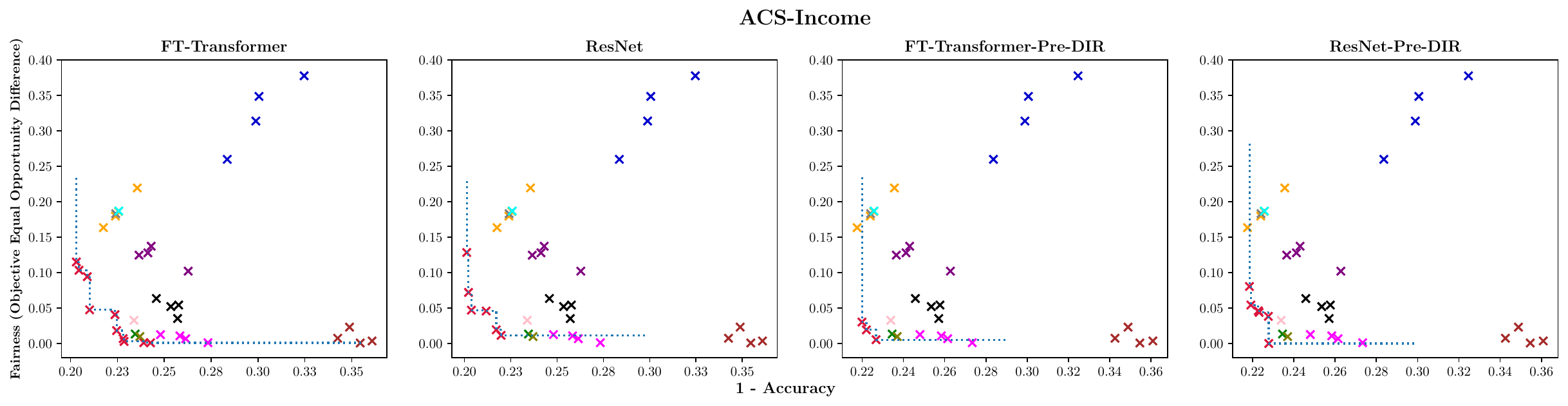}
  % \caption{A subfigure}
  % \label{fig:sub1}
\end{subfigure}

\begin{subfigure}{1\textwidth}
  \centering
  \includegraphics[width=1\linewidth]{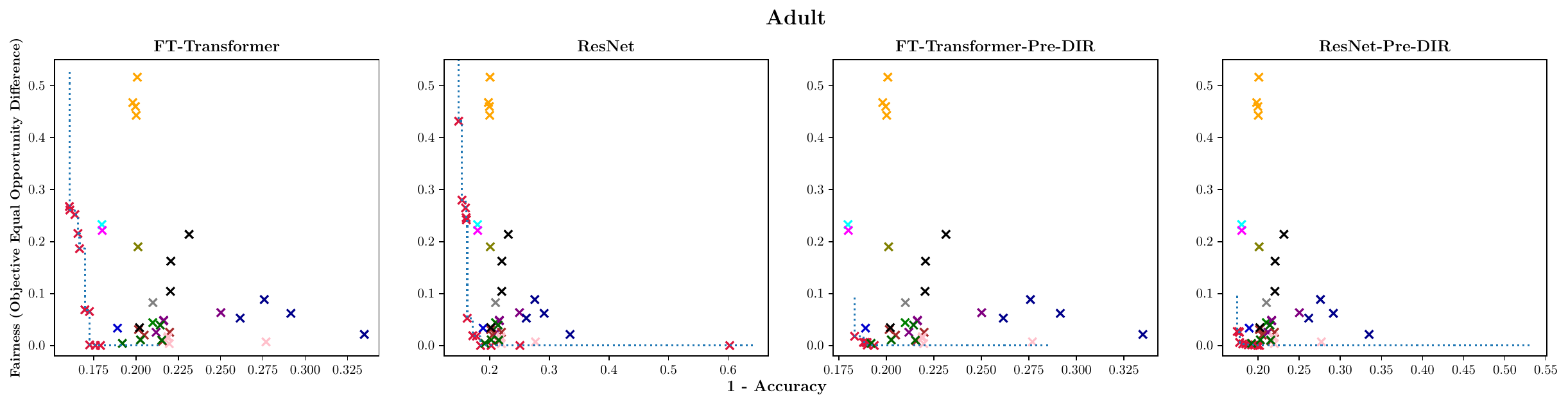}
  % \caption{A subfigure}
  % \label{fig:sub2}
\end{subfigure}

\begin{subfigure}{1\textwidth}
  \centering
  \includegraphics[width=1\linewidth]{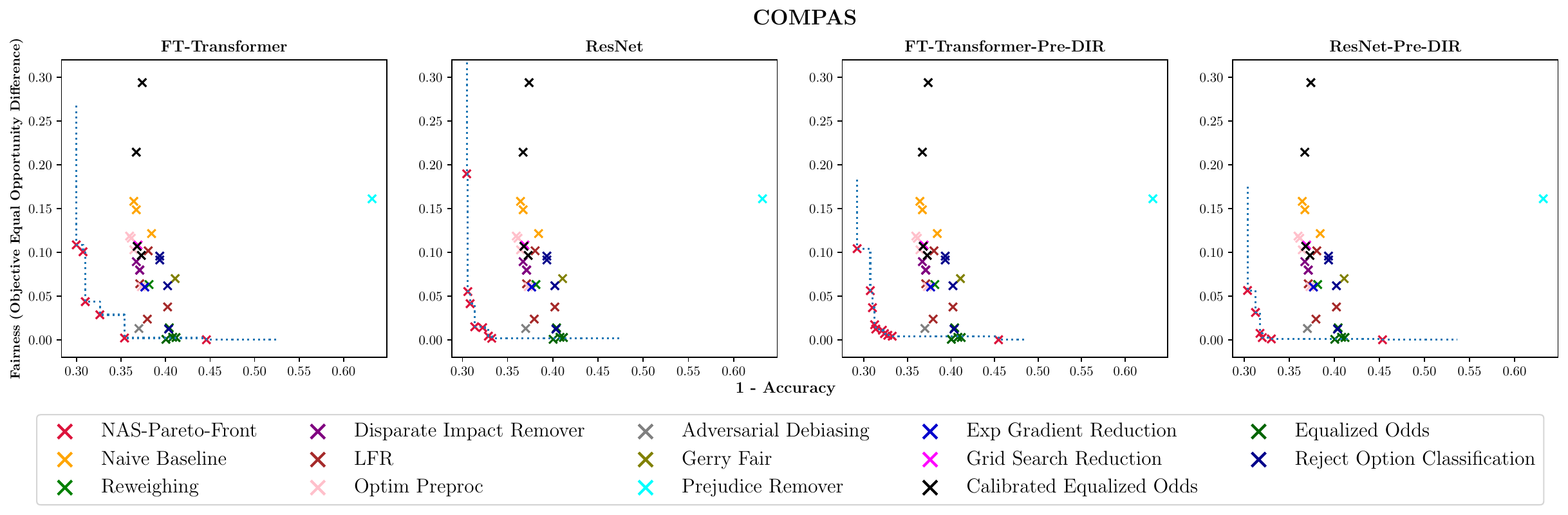}
  % \caption{A subfigure}
  % \label{fig:sub2}
\end{subfigure}
\caption{Pareto fronts for model and pipeline hyperparameters optimized for statistical parity difference and accuracy. The models discovered by \texttt{SMAC} black-box multi-objective optimization are Pareto-optimal on the equal opportunity difference and accuracy metrics compared to a suite of 12 other bias mitigation baselines.}
\label{fig:pareto}
\end{figure}

\paragraph{Multi-objective NAS+HPO.} In Figure~\ref{fig:pareto}, we show the results for our multi-objective optimization runs as outlined in Section~\ref{ssec:optstrat}. Here, we choose the absolute value of statistical parity difference and balanced accuracy to be our optimization targets. We observe Pareto-dominance on almost all the search space configurations over all three benchmarks. The model hyperparameters found on the Pareto front vastly outperform models with default hyperparameters in terms of all the accuracy and fairness metrics (including disparate impact, statistical parity difference, average odds difference, and equal opportunity difference). Different hyperparameters are achieved when optimized with different fairness objectives in mind -- each of them being Pareto-dominant in their trained or related metric. 
In this result of Figure~\ref{fig:pareto}, our NAS+HPO optimization strategy delivers architectures and hyperparameters that outperform all other bias-mitigation methods across standard accuracy, statistical parity difference, and equal opportunity difference. 

We also explore the concept of bias mitigation chaining, where we apply a bias mitigation pre-processing technique and then apply other bias mitigations, including our NAS+HPO approach.
We see however, that our NAS+HPO approach still outperforms baselines in this scenario.
In Figure~\ref{fig:pareto}, the plots labeled as FT-Transformer-Pre-DIR and ResNet-Pre-DIR utilize the Disparate Impact Remover as the pre-processing algorithm before passing it to the neural network classifier. The Disparate Impact Remover makes this pipeline take a hit in accuracy (for instance, in the ACS-Income benchmark) while making these models Pareto-dominant additionally on the disparate impact fairness metric. In these cases, we treat the whole pipeline as a black box and optimize it over the target fairness and accuracy metrics. Optimizing the pipeline as a whole, this strategy helps to search for architectures and hyperparameters that mitigate the accuracy hit taken by the baseline Disparate Impact Remover, and we see how our NAS+HPO approach yields superior results.

\begin{figure}
\centering
\begin{subfigure}{1\textwidth}
  \centering
  \includegraphics[width=1\linewidth]{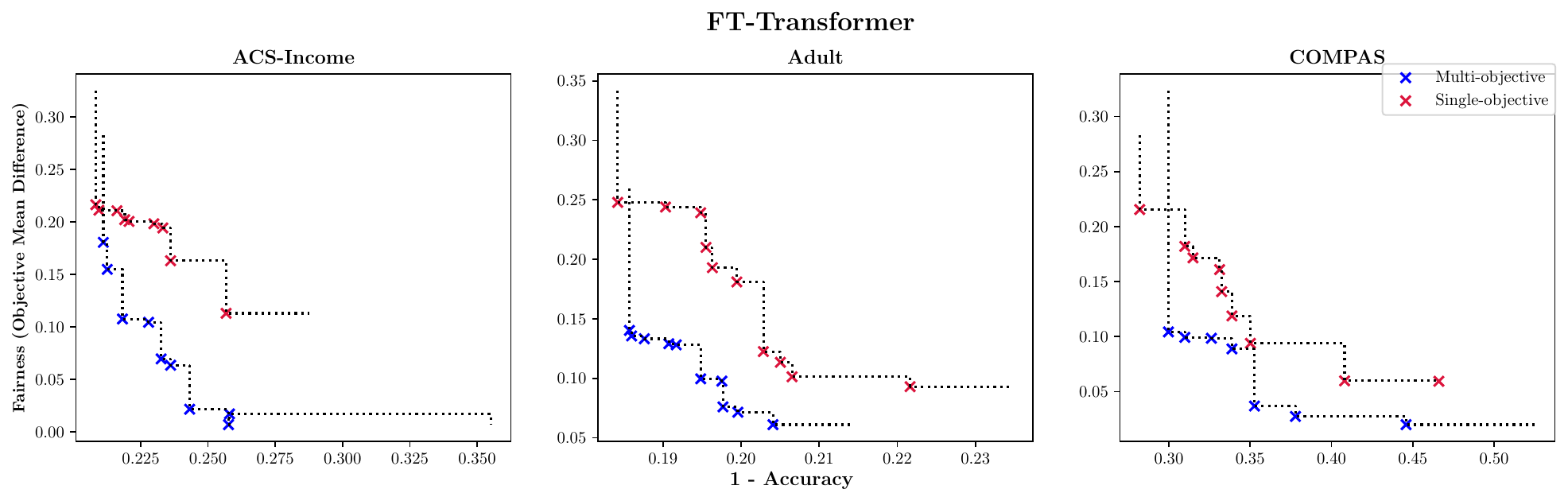}
  % \caption{A subfigure}
  % \label{fig:sub1}
\end{subfigure}

\begin{subfigure}{1\textwidth}
  \centering
  \includegraphics[width=1\linewidth]{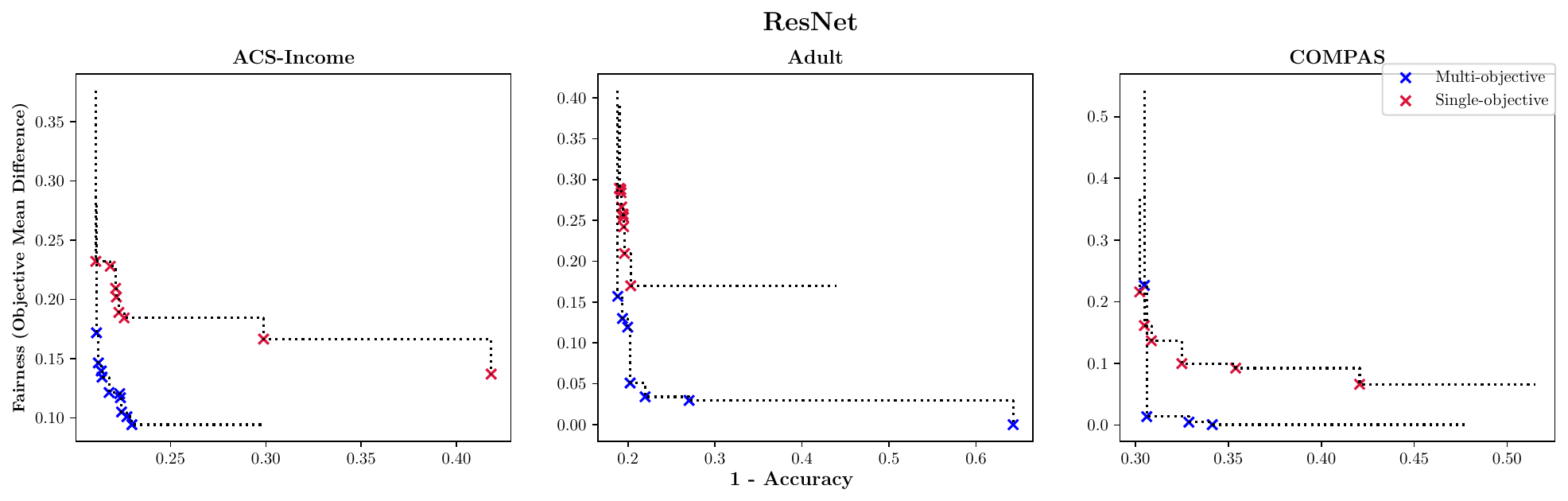}
  % \caption{A subfigure}
  % \label{fig:sub2}
\end{subfigure}
\caption{Single vs Multi-objective Optimization for each search space. Multi-objective NAS+HPO consistently finds models with fairer outcomes in terms of statistical parity difference for each of our benchmark datasets. The single-objective optimization based on the accuracy response sometimes results in models with a bit better predictive accuracy but much worse fairness.}
\label{fig:singlevsmulti}
\end{figure}

\paragraph{Single vs Multi-objective NAS+HPO.} Further, we would also like to understand the trade-off between the fairness and accuracy of the models found by performing NAS+HPO solely based on the accuracy response. We compare the Pareto-fronts obtained from the single and multi-objective optimization runs in Figure~\ref{fig:singlevsmulti}. The models obtained by multi-objective optimization are significantly fairer while taking little to no hit in accuracy. This provides empirical evidence that optimizing the search space without any fairness constraints can be detrimental to the system's overall efficacy in a practical deployment scenario.

\section{Conclusion}

We proposed a novel search space and bias mitigation strategy which combines neural architecture search and hyperparameter optimization. 
Our approach differs from the vast majority of other methods of debiasing and instead uses the inductive bias of an architecture's topology and hyperparameters to try to mitigate forms of algorithmic bias induced by different outcomes in tabular data. 
Specifically, we find that on a wide range of datasets and fairness metrics, our approach produces Pareto-optimal (or often Pareto-dominant) models when compared to extensive existing baselines. 
We started with the observation that our search spaces yielded semantically rich confirgurations which spanned a wide range of accuracy and fairness, i.e., there were exploitable inductive biases in our search space which we beleived NAS and HPO could leverage in search for bias mitigations. 
We leveraged this observation to deploy NAS and HPO algorithms across three common tabular datasets with sensitive attributes. 
Finally, we discovered that our multi-objective approach yielded models which were fairer and more accurate than existing approaches and validated our findings against other forms of debias chaining wherein one debiasing method is applied before another. 

While very promising, we acknowledge that our work comes with its set of limitations --- paticularly in the realm of socially-aware machine learning development. 
We know that our approach can be seen and interpreted as a techno-centrism which tries to address social problems \emph{caused} by the wide-spread application of technology in consequential decision making arena with even more technology. 
Activists and citizens should have the right to disengage with forms of technological interventions. 
We advocate for the continued study of how the concepts of unfairness that have been codified by mathematical notions in the computer science community translate and impact those actually affected ~\citep{Saha20:Measuring}.
We caution that before using any techno-solutionist system, even like ours, in a real world use case, we should always use a critical lens that demands self reflection, and ideally community engagement, as to whether the solution proposed indeed is beneficial to the community or whether it causes more harms \citet{10.1145/3287560.3287598}. 
This work, we hope, will overcome some shortcomings that other pure bias mitigation techniques have, like the portability trap, and can allow domain and public policy experts to optimize the right metric, or suite thereof, instead of just optimizing accuracy or bias separately.

\bibliographystyle{plainnat}
\bibliography{fairness}

\newpage
\appendix

\section{Additional Results}

\begin{figure}[t!]
\centering
\begin{subfigure}{1\textwidth}
  \centering
  \includegraphics[width=1\linewidth]{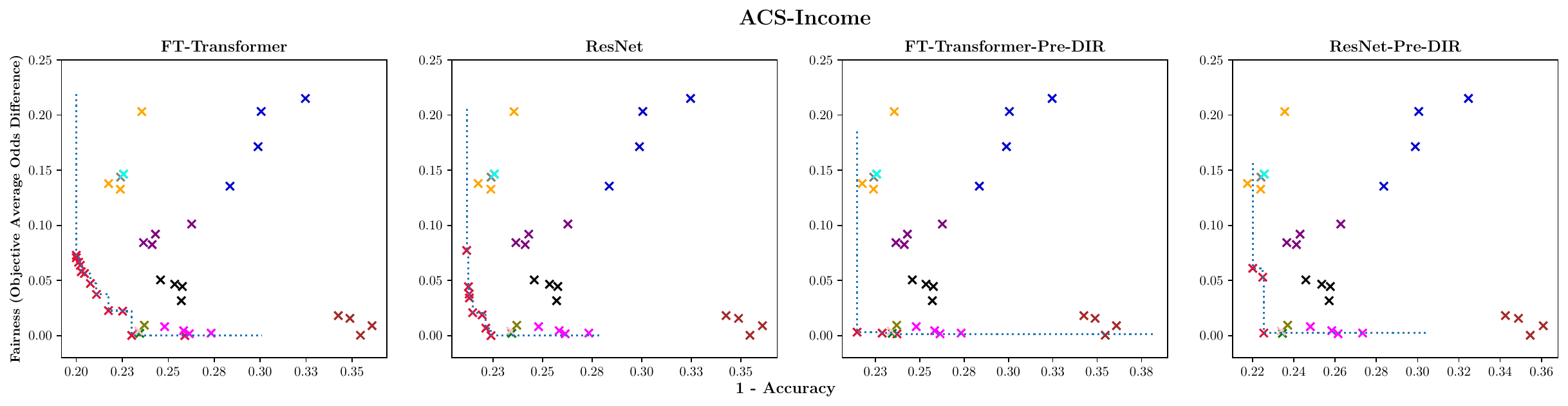}
  % \caption{A subfigure}
  % \label{fig:oddsDiff}
\end{subfigure}

\begin{subfigure}{1\textwidth}
  \centering
  \includegraphics[width=1\linewidth]{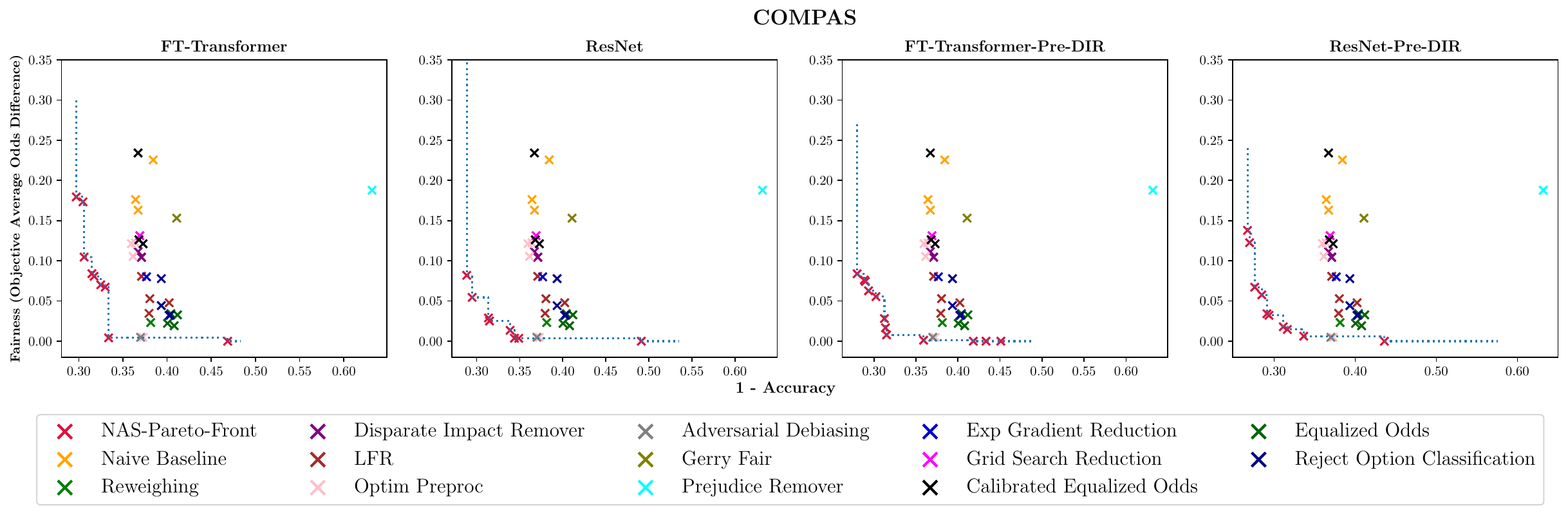}
  % \caption{A subfigure}
  % \label{fig:sub2}
\end{subfigure}
\caption{Pareto fronts for model and pipeline hyperparameters optimized for \textbf{average odds difference} and accuracy. The models discovered by \texttt{SMAC} black-box multi-objective optimization are Pareto-optimal on the average odds difference and accuracy metrics.}
\label{fig:oddsDiff}
\end{figure}

\begin{figure}[t!]
\centering
\begin{subfigure}{1\textwidth}
  \centering
  \includegraphics[width=1\linewidth]{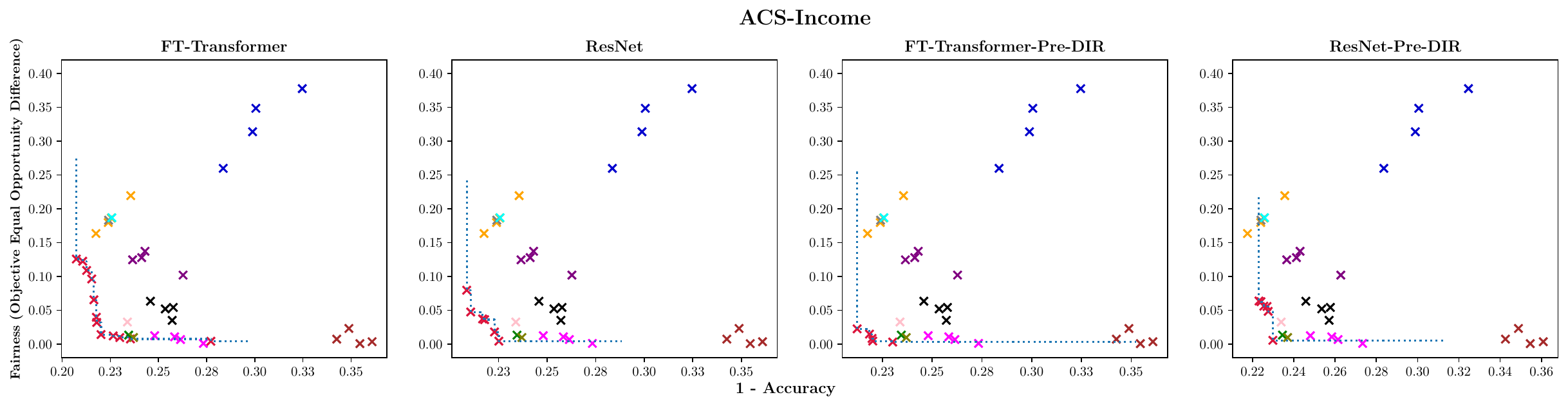}
  % \caption{A subfigure}
  % \label{fig:sub1}
\end{subfigure}

\begin{subfigure}{1\textwidth}
  \centering
  \includegraphics[width=1\linewidth]{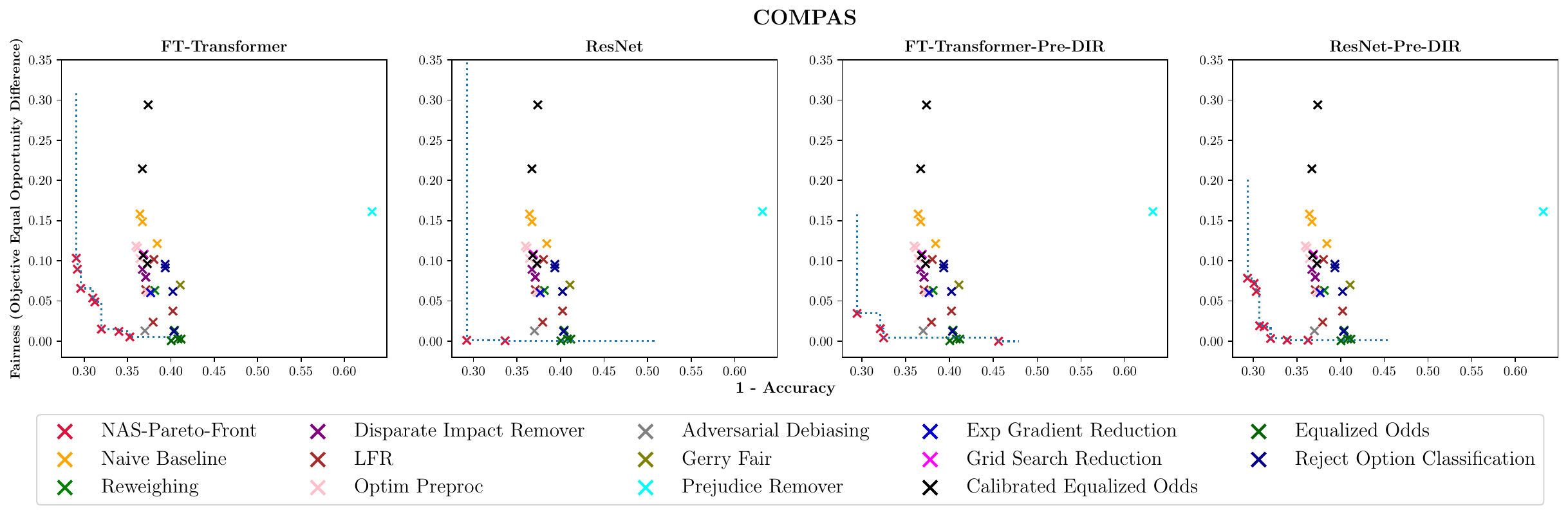}
  % \caption{A subfigure}
  % \label{fig:sub2}
\end{subfigure}
\caption{Pareto fronts for model and pipeline hyperparameters optimized for \textbf{equal opportunity difference} and accuracy. The models discovered by \texttt{SMAC} black-box multi-objective optimization are Pareto-optimal on the equal opportunity difference and accuracy metrics.}
\label{fig:equalOpp}
\end{figure}

We include additional plots that mirror our results in the main portion of the paper but with additional fairness metrics. Specifically, we plot the absolute values of the odds difference in Figure~\ref{fig:oddsDiff} and the equal opportunity difference in Figure~\ref{fig:equalOpp}. Here, we see extensions of our main findings on additional fairness metrics, showing the robustness of our findings and study. 

%%%%%%%%%%%%%%%%%%%%%%%%%%%%%%%%%%%%%%%%%%%%%%%%%%%%%%%%%%%%

\end{document}